\documentclass[conference]{IEEEtran}
 \usepackage[dvips]{graphicx}
\ifCLASSINFOpdf
   \usepackage[pdftex]{graphicx}
\else
   \usepackage[dvips]{graphicx}
\fi

\usepackage{amssymb}
\begin{document}
\title{From Cognitive Binary Logic to Cognitive Intelligent Agents}
\author{\IEEEauthorblockN{N. Popescu-Bodorin*, IEEE Member , and V.E. Balas**, IEEE Senior Member} 
\IEEEauthorblockA{* Dept. of Mathematics and Computer Science, `Spiru Haret' University, Bucharest, Romania\\
 *** Faculty of Engineering, `Aurel Vlaicu' University, Arad, Romania\\
bodorin@ieee.org, balas@drbalas.ro}
}


\maketitle

\begin{abstract}
The relation between 
\textit{self awareness} 
and 
\textit{intelligence} 
is an open problem these days. Despite the fact that 
\textit{self awarness} 
is usually related to 
\textit{Emotional Intelligence}, 
this is not the case here. The problem described in this paper is how to model an agent which
\textit{knows} 
(Cognitive) Binary Logic and which is also able to pass (without any mistake) a certain family of 
\textit{Turing Tests} designed to verify its knowledge and its discourse about the modal states of truth corresponding to well-formed formulae within the language of Propositional Binary Logic. 

\end{abstract}


%
\IEEEpeerreviewmaketitle

\section{Introduction}

The relation between 
\textit{self awareness} 
and 
\textit{intelligence} 
is an open problem these days. Despite the fact that 
\textit{self awarness} 
is usually related to 
\textit{Emotional Intelligence}, 
this is not the case here. The problem described in this paper is how to model an agent which
\textit{knows} 
(Cognitive) Binary Logic \cite{CCBL}  and which is also able to pass (without any mistake) a certain family of 
\textit{Turing Tests} \cite{Turing}  designed to verify its knowledge and its discourse about the modal states of truth \cite{ML} (necessary truth - denoted $t$, contextual truth - denoted $c_{t}$, impossibly truth / necessary false / contradiction - denoted $f$, \cite{CCBL}) corresponding to well-formed formulae within the language of Propositional Binary Logic ($BPL$). 

The context of this paper is given by \cite{synasc} and \cite{CCBL}. More precisely, in order to improve a complex software platform for iris recognition  \cite{synasc}, an inference engine is needed. The computational model of this engine will be derived from Computational formalization of Cognitive Binary Logic ($CCBL$) introduced in \cite{CCBL}. First step in this direction is to extend $CCBL$ up to an intelligent agent enabled to pass some Turing tests, and this is the subject of the present paper.

\section{Preparing for the Turing Test}

In order to pass the Turing test, the agent must have conversational capacities.
Let us assume that the agent gets the input $p$ which \textit{looks} like a well-formed formula of $PBL$. An example of this kind is the \textit{Liar Paradox} discussed in \cite{CCBL}. The problem is that a deductive discourse \cite{CCBL} depends on the given input string but it also depends on the given goal which is obvious for a human agent, but not for a software agent. 

In other words, for a software agent, the input string $p$ can be translated into one of the following sentences: 
 `$p$ is (always) false', 
 `$p$ is (always) true',  
 `$p$ is a contextual truth',  
 `$p$ is false and well-formed',  
 `$p$ is true and well-formed', 
or into one of the following queries: 
 `is it $p$ a theorem?',  
 `is it $p$ a contradiction?',  
 `is it $p$ a contextual truth?'.

\subsection{The cognitive dialect}
\vspace{-4pt}

The introduction of two semantic markers denoted `(!):' and `(?):' is mandatory in order to differentiate between assertions (affirmations) and queries (questions), respectively. With these notations, in the cognitive dialect, the deductive discourses are derived from the deductive discourses written in $CCBL$ by adding `(!):' or `(?):' prefix to each vertices.

The second reason for introducing these markers is that in order to prove a certain degree of self awareness, an agent must be able to understand the difference between the assertions like `I ask' and `I say' and also between `I ask myself', `I say to myself' (`I found', `I proved', `I know'), `I ask you/someone', `I say to you/someone'. 

When it comes to imagining a logical human-machine dialog, the most important thing is that if the human tells something to the agent, then what is told can be true or false, but anything said by the agent must be true (or else, it is inevitably that the agent is inconsistent and, sooner or later, it will fail to pass a certain Turing test). 

Also, to keep the design of our agent as simple as possible, we will consider that all assertions are \textit{positive}, i.e. all of them declare that something is true: 
\\ `it is true that $p$':
\begin{eqnarray}\label{ec-1}
(!): t \rightarrow (p \vee f),
\end{eqnarray}
or `it is true that $p$ is false', 
\begin{eqnarray}\label{ec-2}
(!): t \rightarrow [(p \rightarrow f)\vee f],
\end{eqnarray}
or `it is true that $p$ is a contextual truth':
\begin{eqnarray}\label{ec-3}
(!): t \rightarrow [(c_{t} \rightarrow p)\vee f].
\end{eqnarray}
By analogy, any query will ask for something true: 
\\ `is it $p$ true?':
\begin{eqnarray}\label{ec-4}
(?): t \rightarrow (p \vee f),
\end{eqnarray}
or `is it true that $p$ is false?': 
\begin{eqnarray}\label{ec-5}
(?): t \rightarrow [(p \rightarrow f)\vee f],
\end{eqnarray}
or `is it true that $p$ is a contextual truth?':
\begin{eqnarray}\label{ec-6}
(?): t \rightarrow [(c_{t} \rightarrow p)\vee f].
\end{eqnarray}

The third convention allows the agent to manipulate all three states of modal truth using a purely binary vocabulary. We achieve this by introducing the \textit{dialog function}  (N. Popescu-Bodorin):

\begin{eqnarray}\label{ec-7}
d: \widehat{CI} \rightarrow \{l,n\} \times \hat{t} \cup \hat{f},
\end{eqnarray}
where: $l$ and $n$ are two reserved labels (meaning that the input assertion/query is \textit{logical} or \textit{nonsense}, respectively), $\hat{t}$ is the class of all tautologies, $\hat{f}$ is the class of all contradictions, and $\widehat{CI}$ is defined as follows: if $p$ is a well-formed formula of $PBL$ ($p \in FORM$), then: 
$$[(!): t \rightarrow (p \vee f)]   \in \widehat{CI},$$
$$[(!): t \rightarrow ((p \rightarrow f)\vee f)] \in \widehat{CI},$$
$$[(?): t \rightarrow (p \vee f)]   \in \widehat{CI},$$
$$[(?): t \rightarrow ((p \rightarrow f)\vee f)] \in \widehat{CI}.$$

The output of the  dialog function $d$ is computed using the following rules:
\begin{enumerate}
\item 
 $p$ is a tautology if and only if the full deductive discourse \cite{CCBL} of the  input assertion
 $$(!): t \rightarrow (p \vee f)$$  
 is a deductive proof \cite{CCBL}. In this case, the dialog function outputs the doublet:
 $$(l; t) \in \{l,n\} \times \hat{t}.$$ 
 I.e. the agent proves that $p$ is always true and the input assertion is (logically) well-formed.
\item 
 $p$ is a tautology if and only if the full deductive discourse of the input assertion
 $$(!): t \rightarrow ((p \rightarrow f)\vee f)$$ 
 is a deconstruction \cite{CCBL}. In this case, the dialog function outputs the doublet:
 $$(n; t) \in \{l,n\} \times \hat{t}.$$ 
 I.e. the agent proves that $p$ is always true and founds that asserting falsity for a tautology is a logical nonsense.
\item 
 $p$ is a contradiction if and only if the full deductive discourse of the  input assertion
 $$(!): t \rightarrow (p \vee f)$$
 is a deconstruction. In this case, the dialog function outputs the doublet:
 $$(n;f) \in \{l,n\} \times \hat{f}.$$ 
 I.e. the agent proves that $p$ is always false and founds that asserting truth for a contradiction is a logical nonsense.
\item 
 $p$ is a contradiction if and only if the full deductive discourse of the  input assertion
 $$(!): t \rightarrow ((p \rightarrow f)\vee f)$$ 
 is a deductive proof. In this case, the dialog function outputs the doublet:
 $$(l; f) \in \{l,n\} \times \hat{f}.$$ 
 I.e. the agent proves that $p$ is always false and the input assertion is well-formed.
\item 
 $p$ is a tautology if and only if the full deductive discourse of the input query
 $$(?): t \rightarrow (p \vee f)$$  
 is a deductive proof. In this case, the dialog function outputs the doublet:
 $$(l; t) \in \{l,n\} \times \hat{t}.$$ 
 I.e. the input question is well-formed, the agent proves that $p$ is always true and gives a pozitive answer to the input query.
\item 
 $p$ is a tautology if and only if the full deductive discourse of the input assertion
 $$(?): t \rightarrow ((p \rightarrow f)\vee f)$$ 
 is a deconstruction. In this case, the dialog function outputs the doublet:
 $$(l; f) \in \{l,n\} \times \hat{f}.$$ 
 I.e. the input question is well-formed, the agent proves that $p$ is always true and gives a negative answer to the input query.
\item 
 $p$ is a contradiction if and only if the full deductive discourse of the  input query
 $$(?): t \rightarrow (p \vee f)$$
 is a deconstruction. In this case, the dialog function outputs the doublet:
 $$(l;f) \in \{l,n\} \times \hat{f}.$$ 
 I.e. the input question is well-formed, the agent proves that $p$ is always false and gives a negative answer to the input query.
\item 
 $p$ is a contradiction if and only if the full deductive discourse of the  input query
 $$(?): t \rightarrow ((p \rightarrow f)\vee f)$$ 
 is a deductive proof. In this case, the dialog function outputs the doublet:
 $$(l; t) \in \{l,n\} \times \hat{t}.$$ 
 I.e. the input question is well-formed, the agent proves that $p$ is always false and gives a pozitive answer to the input query.
\item
 $p$ is a contextual truth if and only if none of the full deductive discourses of the input assertions
 $$(!): t \rightarrow (p \vee f)$$
 $$(!): t \rightarrow ((p \rightarrow f)\vee f)$$
or of the input queries 
 $$(?): t \rightarrow (p \vee f)$$
 $$(?): t \rightarrow ((p \rightarrow f)\vee f)$$
 is a deductive proof. In this case, the dialog function outputs the doublet:
 $$(l; [p_{t} \rightarrow (t \rightarrow p)]\wedge[p_{f} \rightarrow (p \rightarrow f)]) $$ 
 I.e. the agent find a context $p_{t}$ which makes the formula $(t \rightarrow p)$ satisfiable and also finds a context $p_{f}$ which makes the formula $(p \rightarrow f)$ satisfiable.

\end{enumerate}

By analyzing the outputs of the \textit{dialog function} it can be seen that, in the \textit{cognitive dialect} is legal to ask \textit{anything} but it is illegal to assert falsity for a tautology or to assert truth for a contradiction.  

\subsection{Simple examples}

Let us consider that the formula $\alpha=(p \rightarrow q)$ is given to be studied.
If the input query is $(?):t \rightarrow (\alpha \vee f)$, a full deductive discourse written in \textit{cognitive dialect} would be:
$$ (?):t \rightarrow ((p \rightarrow q) \vee f) $$
$$ (?):(t \wedge p) \rightarrow (q \vee f) $$
The context which makes the formula $(t \rightarrow \alpha)$ satisfiable is: 
$$\alpha_{t}=[(t \leftrightarrow q)\vee(p \leftrightarrow q)\vee(p \leftrightarrow f)].$$

If the input query is $(?):t \rightarrow ((\alpha \rightarrow f)\vee f)$, a full deductive discourse written in \textit{cognitive dialect} would be:
$$ (?):t \rightarrow (((p \rightarrow q) \rightarrow f)\vee f) $$
$$ (?):(t \wedge (p \rightarrow q)) \rightarrow f $$
$$ (?):[t \rightarrow (p \vee f)] \wedge (?):[(t \wedge q) \rightarrow f] $$
The context which makes the formula $(\alpha \rightarrow f)$ satisfiable is: 
$$\alpha_{f}=[(t \leftrightarrow p)\wedge(q \leftrightarrow f)].$$
The output of the \textit{dialog function} is the following doublet:
 $$(l; [\alpha_{t} \rightarrow (t \rightarrow \alpha)]\wedge[\alpha_{f} \rightarrow (\alpha \rightarrow f)])$$
Hence, $\alpha$ is a contextual truth. Also, $\alpha_{t}$ and $\alpha_{f}$ describe the solutions of Boolean satisfiability problems $(t \rightarrow \alpha)$ and $(\alpha \rightarrow f)$, respectively.

\begin{figure}[!ht]
\centering
\includegraphics[width=2.75in]{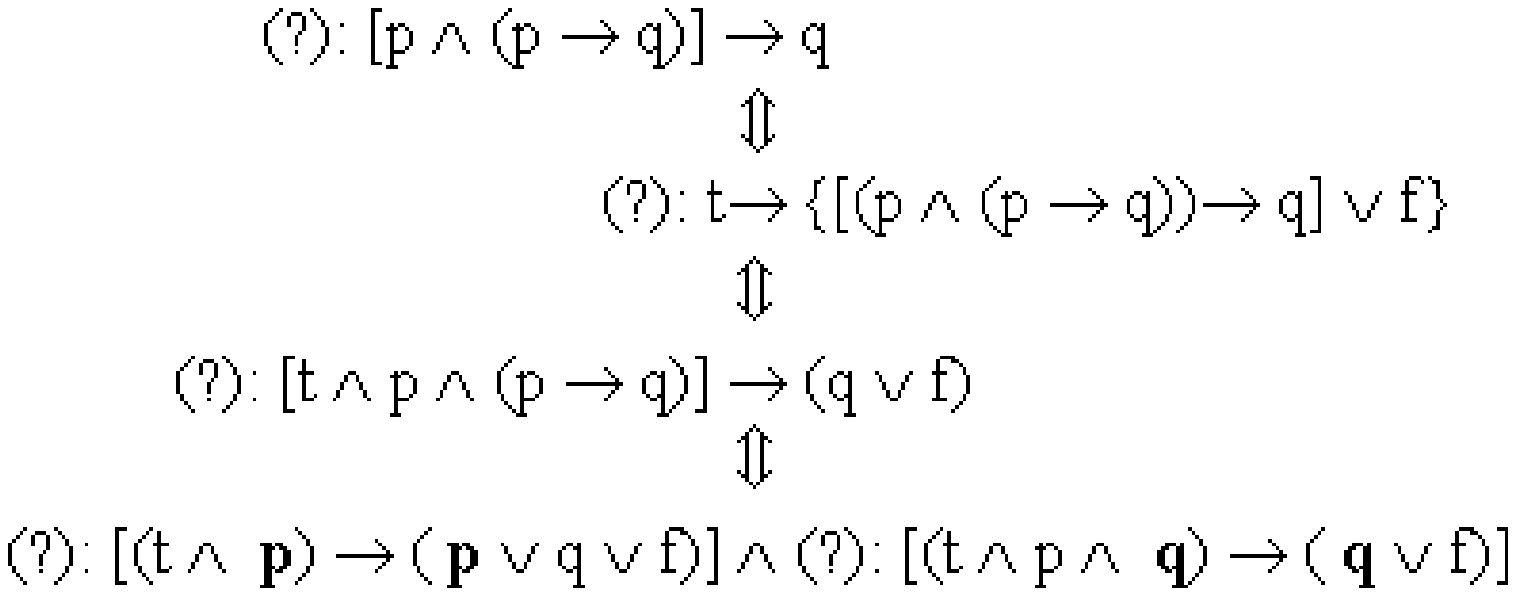}
\caption{A deductive discourse for Modus Ponens written in the \textit{cognitive dialect}.}
\label{MP}
\end{figure}
\begin{figure*}[!t]
\centering
\includegraphics[width=5.55in]{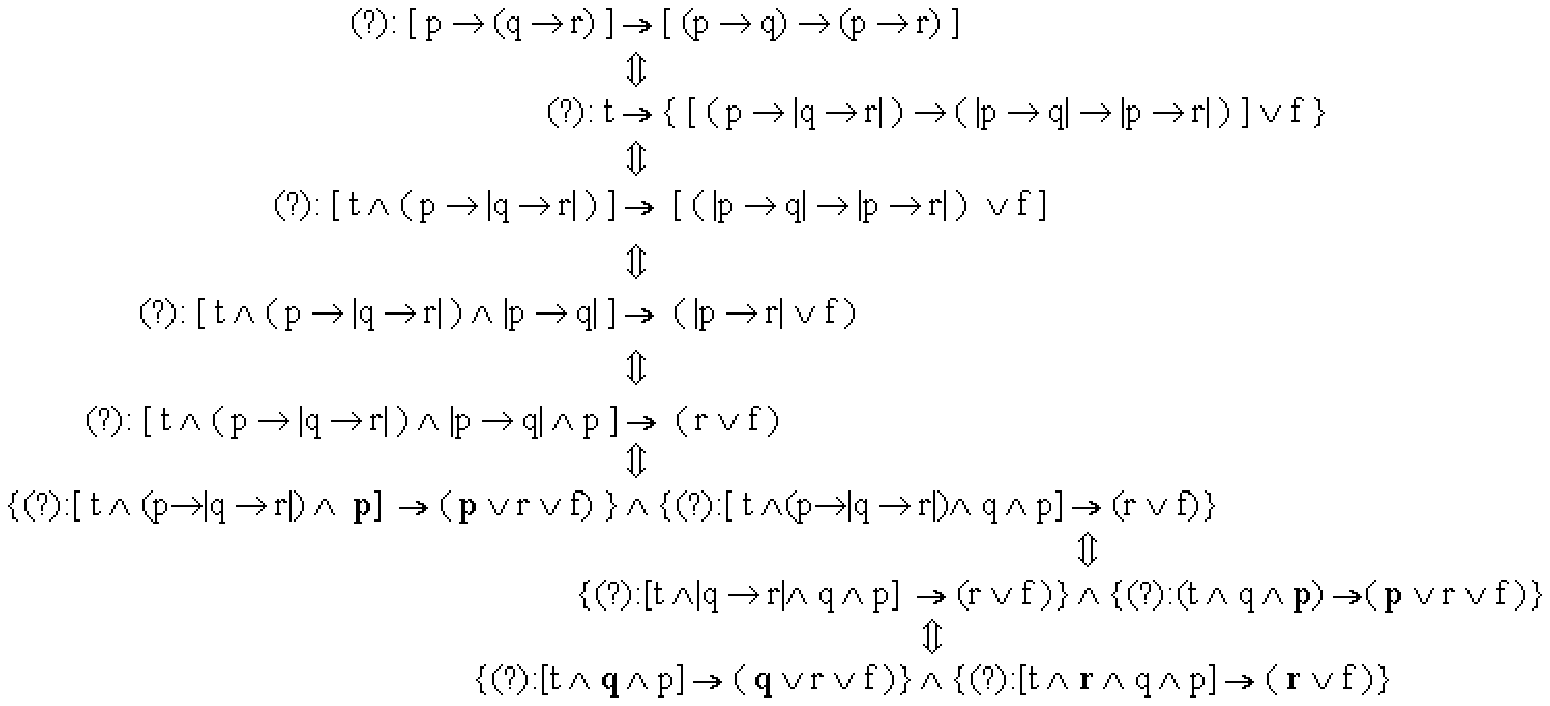}
\caption{A deductive discourse written in \textit{cognitive dialect} for the tautology $[ p \rightarrow (q \rightarrow r) ]{\bf \rightarrow} [ (p \rightarrow q) \rightarrow (p \rightarrow r) ]$.}
\label{LK}
\end{figure*}

In the second example, Modus Ponens is analyzed. For the input query: $$(?): [p \wedge (p \rightarrow q)] \rightarrow q,$$ a full deductive discourse written in \textit{cognitive dialect} would be the deductive proof presented in Fig.\ref{MP} and the output of the \textit{dialog function} is the doublet $(l,t)$.\\

In the third example, the contradiction $(p \wedge \neg p)$ is analyzed. For the input query: 
$$(?): t \rightarrow [((p \wedge \neg p) \rightarrow f)\vee f],$$ a full deductive discourse written in \textit{cognitive dialect} would be the following deductive proof: 
$$(?): t \rightarrow [((p \wedge \neg p) \rightarrow f)\vee f],$$
$$(?): (t \wedge p \wedge \neg p) \rightarrow f $$
$$(?): (t \wedge p) \rightarrow (p \vee f) $$
The output of the \textit{dialog function} is the doublet $(l,t)$. Hence, the input assertion $(!): t \rightarrow [(p \wedge \neg p) \vee f]$
will be recognized as a logical nonsense and the output of the \textit{dialog function} will be the doublet $(n,f)$.\\

The forth example analyze the Modus Tollens argument. For the input query: 
$$(?): (|p \rightarrow q|\wedge|\neg q|)\rightarrow |\neg p|  $$
a full deductive discourse written in \textit{cognitive dialect} would be the following deductive proof: 
$$(?):  t \rightarrow \{[(|p \rightarrow q|\wedge|\neg q|)\rightarrow |\neg p|]\vee f\}$$
$$(?): (t \wedge |p \rightarrow q| \wedge |\neg q| ) \rightarrow (|\neg p|\vee f) $$
$$(?): (t \wedge |p \rightarrow q|\wedge p) \rightarrow (q \vee f) $$
$$[(?): (t \wedge p) \rightarrow (p \vee q \vee f)] \wedge [(?): (t \wedge q \wedge p) \rightarrow (q \vee f)]$$
Hence, the output of the \textit{dialog function} is the doublet $(l,t)$.\\

In the fifth example, we consider the tautology: $$[ p \rightarrow (q \rightarrow r) ]{\bf \rightarrow} [ (p \rightarrow q) \rightarrow (p \rightarrow r) ].$$ 
A deductive discourse written in \textit{cognitive dialect} is presented in Fig.\ref{LK} and the output of the \textit{dialog function} is the doublet $(l,t)$.

\section{The Agent}

\begin{figure}[!ht]
\centering
\includegraphics[width=2in]{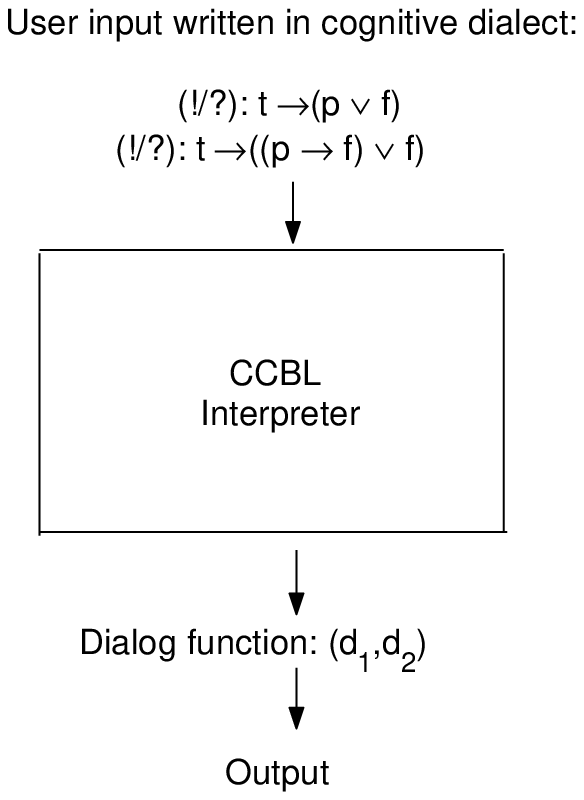}
\caption{The Cognitive Intelligent Agent}
\label{CIA}
\end{figure}

The basic functionality of the proposed Cognitive Intelligent Agent ($CIA$) is described in Fig.\ref{CIA}. Let us consider the Turing tests containing the following type of problems: for an arbitrary formula $p \in FORM$, $CIA$ is required to find if the input assertion/query is or isn't a logical nonsense, and also if $p$ is a tautology, a contextual truth, or a contradiction. 

Since the $CCBL$ theory is sound and complete \cite{CCBL}, $CIA$ will give the correct answer for any input query written in cognitive dialect. Also, if the input assertion is a logical nonsense, $CIA$ will correctly recognize it. Therefore, $CIA$ will pass with a success rate of 100\% all sessions of Turing tests designed to verify its knowledge and its discourse about the modal states of truth corresponding to formulae within the language of $PBL$. Hence, there is no doubt that as a software agent, $CIA$ demonstrate the highest possible degree of intelligence. Still, the agent is not enabled to be aware of itself and of its environment or to simulate self-awareness. 

\section{Conclusion and Future Work}
\label{S3CONCLUSION}

The basic design of an inteligent agent was proposed in this paper. It is an example of a fully intelligent agent which is not at all aware of itself. Still, it is enabled to engage in simple conversations about the modal states of truth of well-formed formulae of $PBL$, without doing any mistakes. 

Future developments will include self-awarnes which is needed in order to enable the agent to supervize complex computations and to engage direct communication with humans on specific subjects, other than the modal truth state of the formulae of PBL. For example, we plan to gather humans` opinions about some particular iris recognition results in a similar manner with that described in \cite{WSEAS1} where customers` emotive responses to a product are collected using a questionnaire. The goal is corelate iris recognition results obtained automaticaly with human feedback and to explore the limitations that could apear in iris recognition when using eye images captured in insufficiently constrained aquisition conditions.

On the other hand, we know that some sub-problems of iris recognition are in NP (the class of problems solvable in nondeterministic polynomial time), and consequently heuristic algorithms must be used in order to achieve some speed. The problem is that quantifying the quality of their results is a \textit{matter of degree} \cite{Zadeh1} (in fact, the results of these heuristic algorithms are \textit{near solutions}, not exact solutions). Therefore, future developments in the direction of fuzzy logic are not excluded at all.

\end{document}